\documentclass[10pt,twocolumn,letterpaper]{article}

\usepackage{wacv}
\usepackage{times}
\usepackage{epsfig}
\usepackage{graphicx}
\usepackage{amsmath}
\usepackage{amssymb}
\usepackage[english]{babel}
\usepackage{subfig,color}
\usepackage{float}

\newcommand{\minisection}[1]{\vspace{0.04in} \noindent {\bf #1}\ \ }

\wacvfinalcopy 

\def\wacvPaperID{****} 

\ifwacvfinal\fi
\begin{document}

\title{Bandwidth limited object recognition in high resolution imagery}

\author{Laura Lopez-Fuentes \\
AnsuR Technologies, Norway\\ Computer Vision Center, UAB, Spain\\ Dept. Mathematics and Computer Science, UIB, Spain\\
{\tt\small laura@ansur.no}
\and
Andrew D.Bagdanov \\
University of Florence, Italy\\
{\tt\small andrew.bagdanov@unifi.it}
\and
Joost van de Weijer \\
Computer Vision Center, UAB, Spain\\
{\tt\small joost@cvc.uab.es}
\and
Harald Skinnemoen\\
AnsuR Technologies, Norway\\
{\tt\small harald@ansur.no}
}
\maketitle
\ifwacvfinal\thispagestyle{empty}\fi

\begin{abstract}
This paper proposes a novel method to optimize bandwidth usage for
  object detection in critical communication scenarios. We develop two
  operating models of active information seeking. The first model
  identifies promising regions in low resolution imagery and
  progressively requests higher resolution regions on which to perform
  recognition of higher semantic quality. The second model identifies
  promising regions in low resolution imagery while simultaneously
  predicting the approximate location of the object of higher semantic
  quality. From this general framework, we develop a car recognition
  system via identification of its license plate and evaluate the
  performance of both models on a car dataset that we introduce. Results
  are compared with traditional JPEG compression and demonstrate that
  our system saves up to one order of magnitude of bandwidth while
  sacrificing little in terms of recognition performance.
\end{abstract}

\section{Introduction}
\label{sec:introduction}
\begin{figure}
\centerline{\includegraphics[width = \linewidth]{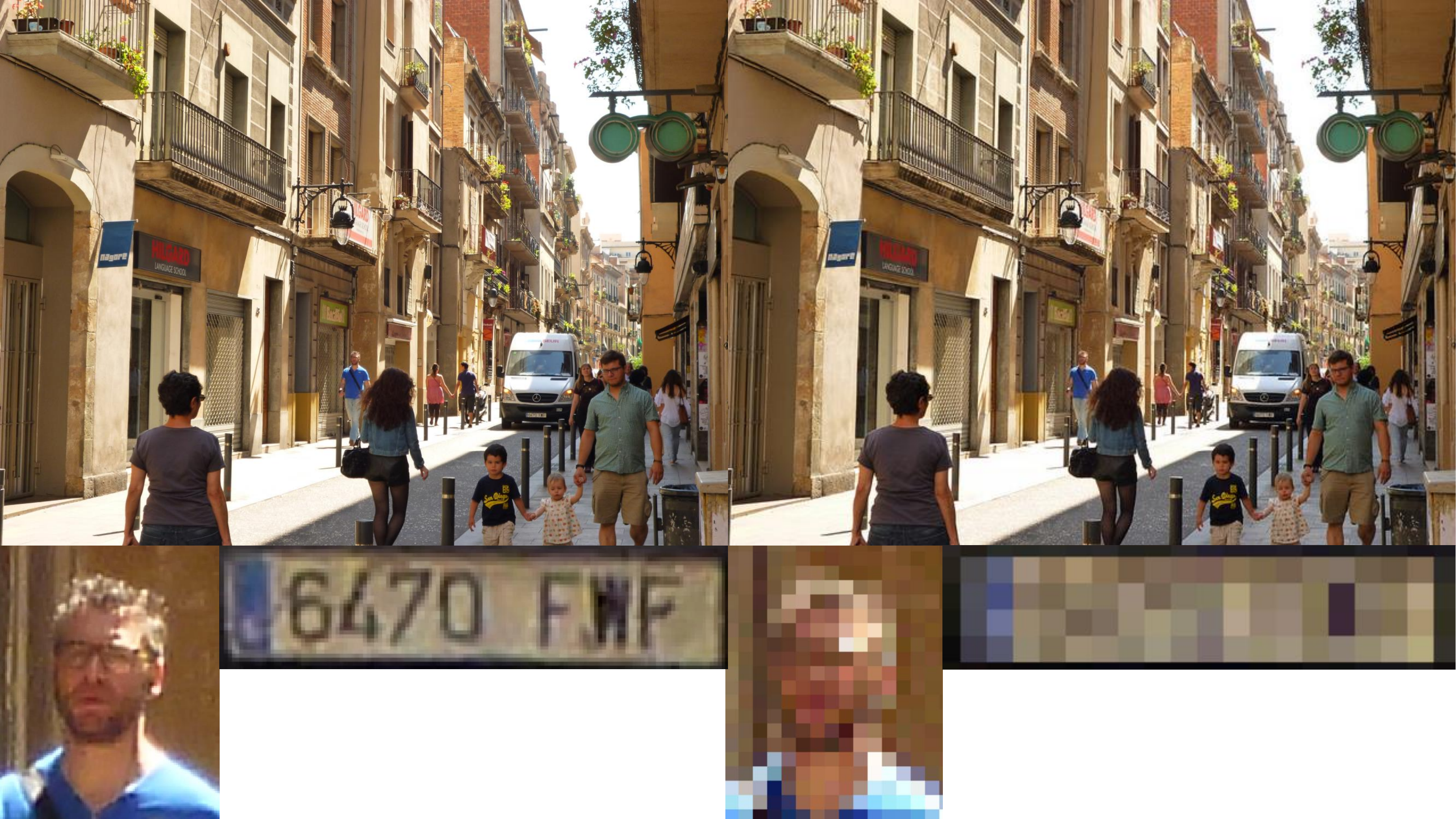}}
\caption{The top-left image corresponds to the original 16 megapixel
  image, and the top-right image to a downscaled version at
  640$\times$480 pixels. The information loss may not be visible at
  first glance, however when zooming in it is clear that valuable
  information such as faces or license plate numbers have been lost.}
\label{fig:introduction}
\end{figure}

The explosion in mobile phone ownership has made taking and sharing
pictures commonplace in our daily lives. Average smartphones capture
10 megapixel images, and high-end smartphones capture up to 20
megapixel images. Such high resolution images provide detailed
information on the objects in the scene, which are only appreciated
after zooming into these regions. For example, Figure
\ref{fig:introduction} shows the same image at two different
resolutions. On the left, the upper image corresponds to the original
image with its original 16 megapixel resolution. On the right, the top
image corresponds a downscaled version of the image -- in this case
640$\times$480 pixels (0.30 Mpixel). This is a common resolution used in
computer vision and almost the highest resolution that can be fed into
state-of-the-art convolutional neural networks. At first glance both
images look fairly similar. However when zooming in, the original
resolution image preserves information valuable for performing tasks
such as face recognition or license plate transcription, while the
lower resolution image has completely lost that information. In this
article we investigate the detection of objects in high resolution
imagery for bandwidth critical applications. We focus on objects
which are structurally related to others (like license plates with
cars, and faces with persons).

Low bandwidth communication is important for all applications where
computation is performed at the server-side. For such applications reducing
the amount of transmitted data is crucial for the speed of the
application and to reduce costs. This is especially true when
considering satellite communications which are used extensively in
situations where standard phone networks are absent (e.g. fire
detection with drones), in emergency situations when there is a sharp
increase in call volume and terrestrial network capacity is saturated
(or absent due to a disaster) or in high security situations where
satellite communications is more secure~\cite{pimentel2012optimized,skinnemoen2012asign}.  The objective of this work is to improve the efficiency of visual communication in bandwidth-limited scenarios. Our emphasis is on conserving total pixels inspected (and thus transmitted), rather on end-to-end computational efficiency. Especially in satellite communications, where bandwidth costs can be as high as \$1 per megabyte, it is more critical to reduce bandwidth consumption and rather than optimizing computational efficiency.


Image compression is a standard approach in the communications field
when it is necessary to cope with bandwidth limitations and
costs. These techniques exploit the highly correlated nature of
natural images~\cite{wallace1991jpeg, weinberger2000loco}. However,
this practice, besides reducing the bandwidth needed to send the
image, also reduces the information contained, which in some cases may
result in critical information losses. To obtain superior bandwidth
reduction, some applications resort to human operators who
interactively and progressively request higher resolution of regions
of interest from a low resolution image. As such, they discard large
parts of the image which are considered redundant for the task at
hand. In this paper we propose a method for Multi-stage object
detection in which we maintain the computation and bandwidth benefits
of working with low resolution images while keeping the valuable
information that is contained in the original image.

Object recognition has seen significant changes over the last
decade. It has long been dominated by sliding window
approaches~\cite{dalal2005histograms, felzenszwalb2010object}. An
alternative approach is based on object proposals which reduce the
number of windows considerably thereby allowing the use of more
complex classifiers~\cite{krahenbuhl2014geodesic, uijlings2013selective, zitnick2014edge}. The breakthrough in deep convolutional
networks, which first showed remarkable results on image
classification~\cite{krizhevsky2012imagenet}, was almost immediately
extended to object proposal methods, and has resulted in a
considerable performance improvement for object
detection~\cite{girshick2014rich}. Several papers have further
improved these results and increased speed~\cite{girshick2015fast,
  redmon2015you}. As a result of these improvements the accuracy of
object detection greatly improved, and these techniques can now be
used as reliable building blocks in computer vision pipelines.

Actually, detection in high resolution images has received relatively
little attention -- the fast growth of image size in commercial
cameras seems to have gone unnoticed by the object detection
community. Most research has focused on detecting relatively large
objects in low resolution images. In fact, popular state-of-the-art
object detection algorithms based on convolutional neural networks
(CNN) cannot handle high resolution imagery, for example, the popular
AlexNet \cite{krizhevsky2012imagenet} rescales all images to a meager
224$\times$224 pixels (0.05 megapixels) before processing. As
discussed before this resolution reduction comes at the cost of losing
valuable information, which in turn results in an inability to detect or recognize
all objects present in the image. 

The image interpretation process needed for interactive, low bandwidth
image communication is traditionally done by humans, which is costly
and slow. We propose to apply object recognition to automatically
identify the relevant regions in images. We call our approach
\emph{active information seeking} because it mimics the actions a
human operator would perform in order to actively identify semantic
objects while consciously and actively limiting the bandwidth
consumed. We investigate two approaches for the active information
seeking for objects which are structurally related to others (like
license plates with cars, and faces with persons). Firstly, we propose
a Multi-stage approach which identifies promising regions of interest
in low resolution imagery and progressively requests the regions in
higher resolutions to perform recognition of higher semantic
quality. Secondly, we propose a direct-estimation approach which
detects objects (e.g. cars) in images and directly estimates the location of their
parts (e.g. the license plate). This allows to directly extract the
part at a higher resolution, thereby further reducing bandwidth usage.


This paper is organized as follows. In the next section we describe a
framework for Multi-stage recognition models that allows us to
quantify bandwidth savings of one model versus another. In
Section~\ref{sec:approach} we present our approaches for
bandwidth-limited object recognition. Then, in
Section~\ref{sec:experiments} we report on a number of experiments we
performed to quantify the performance of our approach with respect to
baselines and the state-of-the-art. Finally, we conclude with a
discussion of our contribution in Section~\ref{sec:conclusions}.

\section{A framework for bandwidth-limited recognition}
\label{sec:method}
In this section we describe our architecture for bandwidth-limited
recognition in high resolution images. Our approach is based on
actively selecting which parts of the image to inspect at higher
resolutions. This allows our system to save bandwidth by ignoring
semantically irrelevant portions of images. The model takes a low
resolution image from which consecutive requests of higher resolution
regions are made to extract the desired information. This model can be
applied to detect semantic objects contained in a sequence of objects
of decreasing size. For example, the license plate is contained in the image of the car,
which is itself contained in an image of a street scene. With our
approach, we can access high resolution images of such objects without
having to inspect the entire image at high resolution. This model can be applied to an interactive visual communication system for bandwidth limited channels where images are first sent in low resolution and then parts of it can be requested in higher resolution, this allows large amounts of savings in bandwidth with no sacrifice on details which may be relevant. 

\begin{figure}[t]
\centerline{\includegraphics[width = 0.8\linewidth]{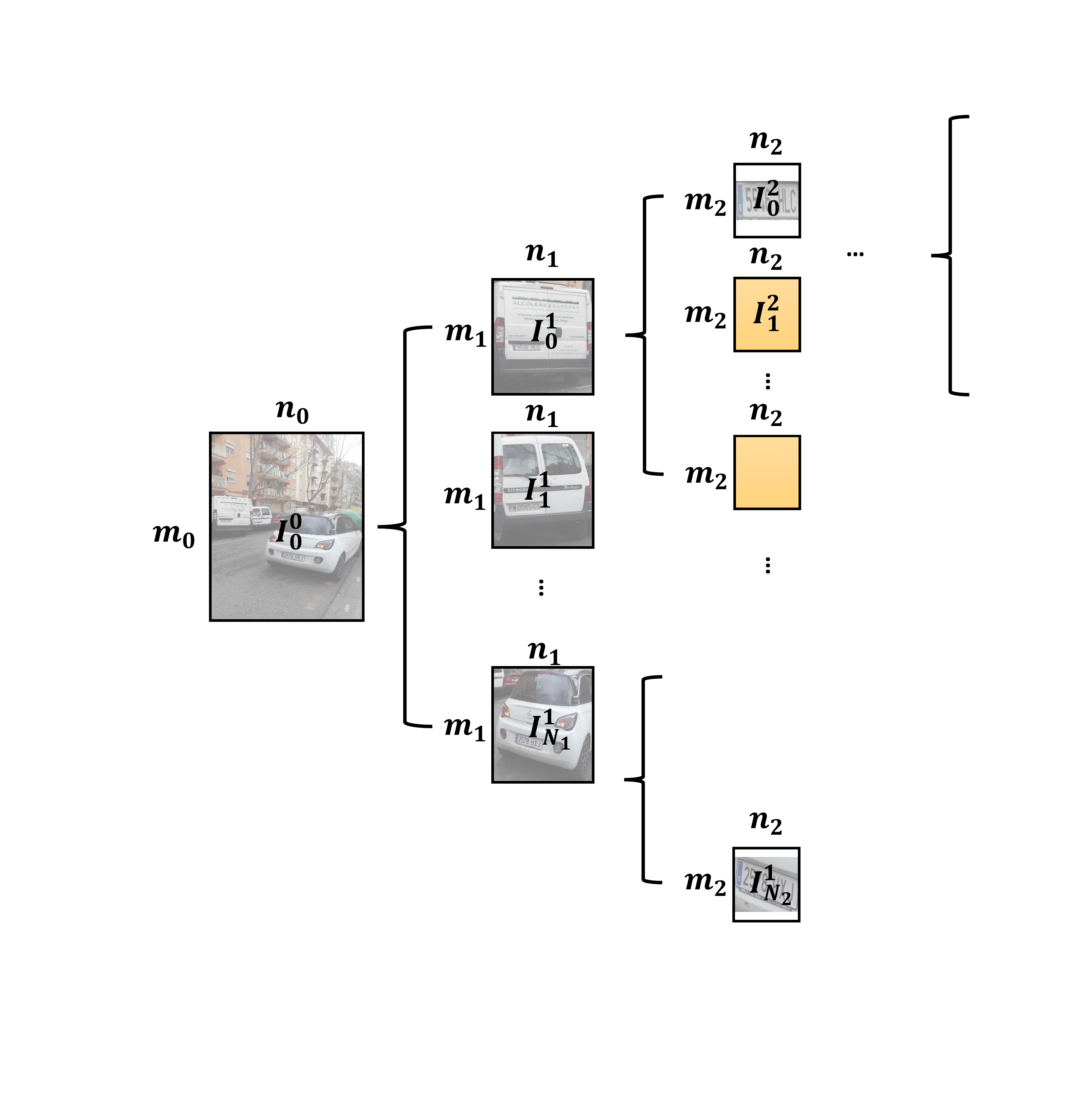}}
\caption{A graphical overview of our approach. A low resolution
  version of the original image is first inspected to identify regions
  of interest (e.g. cars if we are searching for license
  plates). These regions are then inspected at higher resolution to
  search for objects of higher semantic interest (e.g. license plates
  themselves).
}
\label{fig:model}
\end{figure}

Our object recognition framework is illustrated in
Figure~\ref{fig:model}. We use a Recursive model of information
seeking that identifies promising regions to inspect at increasingly
higher resolutions. In this figure, $I^i_j$ stands for the $j$-th
subimage inspected at the $i$-th level of resolution. Subimages at
level $i$ are all of the same size of $n_i \times m_i$ pixels. At each
level $i$ of the model, $N_i$ subimages are inspected at $n_i \times
m_i$ pixel resolution.

The system begins with the original image $I^0_0$ at low resolution
as input. A detector (or some other method to identify promising
subimages) is then applied. The detector detects $N_1$ regions of
interest, e.g. cars if our final goal is to identify license plates. Each of these subimages is then inspected at stage 1 of the
model, and a total of $N_2$ regions of interest are passed to stage 2
for inspection at higher resolution. This process is repeated $n$ times,
where $n$ varies depending on the application. Finally, the desired
information is extracted from the last sequence of
subimages.

Following this notation, the cost of inspecting the sequence of images
is:
\begin{equation}
\text{Cost} = \rho \sum_{i=0}^n{N_i \cdot n_i \cdot m_i},
\label{eq:cost}
\end{equation}
where $\rho$ is a constant representing the bandwidth in MB needed to
send one pixel. Our final goal is to have a final cost much lower than
the cost of inspecting the image for the desired objects of interest
at its original high resolution.




\section{Two models for bandwidth limited recognition}
\label{sec:approach}

In this section we propose two different models for bandwidth limited object
recognition. The first is based on Recursive application of the
Fast-RCNN detector~\cite{girshick2015fast}. Then, in
Section~\ref{sec:multi-stage} we describe a second approach based on
simultaneous estimation and localization of regions of interest and
high-resolution objects contained in them.

\subsection{Recursive Fast-RCNN}
\label{sec:fast-rcnn}
To select the regions which will be requested at higher resolution we
use the Fast-RCNN object detector~\cite{girshick2015fast}. Fast-RCNN
was originally proposed to directly detect objects from a single
resolution of an input image.

\minisection{The Fast-RCNN detector.}
Assume there are $C$ object classes we wish to both recognize and
localize in images. A trained Fast-RCNN network takes as input an
image $I$ and a sequence of $R$ bounding box proposals:
\begin{eqnarray*}
B = [ \mathbf{b}_i \ | \ i \in \{1, 2, \ldots, R\} ],
\end{eqnarray*}
where each $\mathbf{b}_i = (x_i, y_i, w_i, h_i)$ encodes the geometry
of the $i$-th bounding box proposal.  These proposals may be generated
by Selective Search~\cite{uijlings2013selective}, Edge
Boxes~\cite{zitnick2014edge}, or any bounding box proposal strategy
one prefers~\cite{hosang2016makes}.

The Fast-RCNN network then produces a structured output ${P}$,
which consists of two elements for each bounding box proposal in image:
$$
  {P} = \left\{ [{p}(c | \mathbf{b}_i), \hat{\mathbf{b}}_{i|c}] \ \ | \ \ i = 1, 2, \ldots, R \right\}.
$$
Each ${p}(c | \mathbf{b}_{i})$ is the estimated probability that
box $\mathbf{b}_i$ belongs to class $c$. Each
$\hat{\mathbf{b}}_{i|c}$, on the other hand, is a \emph{refined
  bounding box prediction} that adjusts the original proposed box
geometry $\mathbf{b}_i$ (using visual information from $I$) to
better predict the expected bounding box location and extent of a box
from class $c$. The final predicted class probabilities and boxes are
computed using standard non-maximum suppression on ${P}$. 

Internally, the predictions of both ${p}(c | \mathbf{b}_i)$ and
$\hat{\mathbf{b}}_{i|c}$ are made on the basis of a visual representation
of each \emph{box} $\mathbf{b}_i$ extracted from neuron activations in
intermediate layers of a CNN (typically the last fully-connected layer
of the network). This is an essential feature of the Fast-RCNN framework:
the convolutional features are computed over the entire \emph{image}
up to a pre-defined point in the network, then these features are
pooled into the representations of each bounding box proposal in
$B$. These are then sent down to the rest of the network which estimates
outputs ${p}(c | \mathbf{b}_i)$ and $\hat{\mathbf{b}}_{i|c}$.

A Fast-RCNN network is trained by optimizing an average per-box
multitask loss over all training images. This multitask loss drives
the network to both recognize \emph{and} localize the desired object
classes by simultaneously learning a classifier for all object
categories and a regressor from bounding box proposals to boxes that
(on the basis of visual content of the proposed bounding box) better
fit the ground truth object box
annotations. See~\cite{girshick2015fast} for complete details of this
optimization procedure.

\minisection{Recursive application of Fast-RCNN.}
In order to actively seek objects at high resolution, as described in
the previous section, we require a Fast-RCNN detector that works at
multiple levels of resolution. In this first model, we achieve this by
applying Fast-RCNN Recursively. In our application, a car detector
based on Fast-RCNN is first run, after which a second Fast-RCNN
network trained to detect license plates is run on the selected car
bounding boxes extracted at high resolution. Finally, an off-the-shelf
OCR engine is used to recognize the text of each localized license
plate.


\subsection{Multi-stage network}
\label{sec:multi-stage}


The main objective of our framework is to perform object recognition
and minimize the bandwidth needed, where bandwidth is measured in the
total number of pixels inspected by the detector. In this section we
propose a second model that saves even more bandwidth than the one
proposed above. In this model, in addition to detecting an object of
interest at each stage, we simultaneously predict the subregion which
is likely to contain the object or objects of interest in the next
stage. Thus, instead of requesting the entire object of interest from
the previous stage (e.g. the entire car detection), we directly
request a smaller region of interest for the next stage (e.g. a smaller region which is expected to contain the license plate). With respect
to the Recursive Fast-RCNN model described in the previous section,
this allows us to skip an entire Fast-RCNN stage and thus save more
bandwidth. In the car recognition example, instead of requesting the
\emph{cars} at a higher resolution, at the second stage of processing
we can directly request the \emph{license plates} at higher resolution
and proceed to the OCR stage.

To implement this Multi-stage model, we propose a novel network
architecture. The proposed network has the same inputs as Fast-RCNN,
an input image $I$ and a sequence of $R$ bounding box
proposals. However, the output ${P}$ produced for each bounding box
proposal, contains \emph{four} elements instead of two:
\begin{eqnarray*}
  {P} = \left\{ [{p}(c | \mathbf{b}_i), \hat{\mathbf{b}}_{i|c}, {p}(s | \mathbf{\hat{b}}_{i|s}), \hat{\mathbf{b}}_{i|s}] \ | \ i
              = 1, 2, \ldots, R \right\},
\end{eqnarray*}
where ${p}(c | \mathbf{b}_{i})$ is the estimated probability that box
$\mathbf{b}_i$ belongs to class $c$, $\hat{\mathbf{b}}_{i|c}$
corresponds to the \emph{refined bounding box prediction} to the main
object, $\hat{\mathbf{b}}_{i|s}$ stands for a second regression done
from the proposed bounding box which corresponds to the localization
of the object which is to be detected in the next stage (the
sub-object).  Finally, ${p}(s | \mathbf{\hat{b}}_{i|s})$, is the
estimated probability that the bounding box from the second regression
belongs to class $s$.

\begin{figure*}
\centerline{\includegraphics[width = 0.75\linewidth]{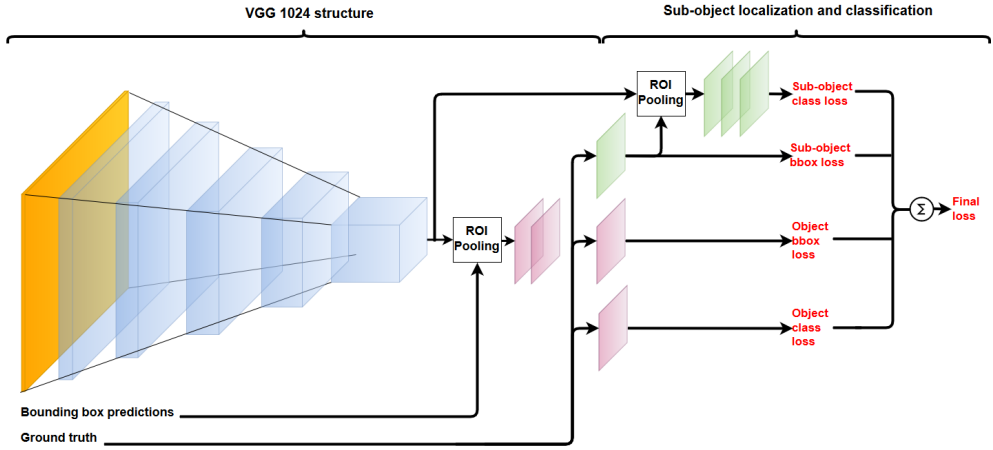}}
\caption{Diagram of the proposed Multi-stage network. The yellow box
  corresponds to the original image fed to the network, the blue boxes
  correspond to convolutional layers, the purple ones to fully
  connected layers, and the red letters to different losses to
  optimize. The proposed network follows a similar structure as
  VGG1024 up to the fully connected layer fc7, from which we have
  added extra layers to enable sub-object detection and localization.}
\label{fig:network}
\end{figure*}

The basic architecture of the network is similar to the Fast-RCNN
network with the two different heads for the object class loss and the
object bounding box loss (see Figure~\ref{fig:network}). In our
Multi-stage network, we introduce another regression head to directly
estimate the coordinates of the sub-object (the object to recognize at
the next stage). In addition, we introduce a classification head to
compute the sub-object classification score which can be used as a
quality measure of the sub-object coordinates.

Directly estimating this sub-object classification loss in parallel
with the other three heads was found to provide sub-optimal
results. Therefore, we introduce an additional pooling layer to
extract the features within the estimated bounding box of the
sub-object. In order to obtain these features from the last
convolutional layer an unnormalization layer is needed. This layer
changes the coordinates of the estimated bounding box from being
normalized with respect to the bounding box proposed by selective
search to being normalized with respect to the whole image. In
Section~\ref{sec:experiments} we show that the output of the
sub-object classification layer can be exploited when deciding on an
additional margin to the bounding box of the sub-object to
extract. When the sub-object classification score is high no
additional margin is required. For low scores a large margin is used
to ensure the sub-object is within the extracted region.

To train the network we minimize the sum of the losses over all ROIs
and all training images. Specifically the loss of a single ROI has
four components:
\begin{eqnarray*}
L_j\left({P_j}, GT_j\right) = L_{\text{obj}}^{\text{cls}} + L_{\text{sub-obj}}^{\text{cls}}  +  L_{\text{obj}}^{\text{bbox}} + L_{\text{sub-obj}}^{\text{bbox}},
\end{eqnarray*}
where ${P_j}$ are the estimated outputs from the j-th ROI, and $GT_{_j }  = \left\{ {c_j^{GT} ,s_j^{GT} ,\mathbf{b}_{j|c}^{GT} ,\mathbf{b}_{j|s }^{GT} } \right\}$ is the ground truth of the object class $c_j^{GT}$, the sub-object class $s_j^{GT}$, and their ground truth bounding boxes, $\mathbf{b}_{j|c }^{GT}$ and $\mathbf{b}_{j|s }^{GT}$. The class loss is given by log loss of the true class:
\begin{eqnarray*}
 L_{\text{obj}}^{cls}  &=&  - \log p( {c_j^{GT} |\mathbf{\hat{b}}_j } ) \\
 L_{\text{sub - obj}}^{cls}  &=&  - \log p( {s_j^{GT} |\mathbf{\hat{b}}_{i|s}} ).
\end{eqnarray*}

The location losses on the bounding boxes are defined as:
\begin{eqnarray*}
L_{\text{obj}}^{\text{bbox}} &=& \left[ {c_j^{GT}  \ge 1} \right] \text{smooth}_{L1}(\hat{\mathbf{b}}_{j|c} - \mathbf{b}_{j|c }^{GT} )\\
L_{\text{sub-obj}}^{\text{bbox}} &=& \left[ {s_j^{GT}  \ge 1} \right]  \text{smooth}_{L1}( \hat{\mathbf{b}}_{j|s} - \mathbf{b}_{j|s }^{GT})\\
\end{eqnarray*}
where the Iverson bracket indicator function $\left[ {c_j^{GT}  \ge 1} \right]$ is one if ${c_j^{GT}  \ge 1}$ and zero otherwise. For bounding boxes which do not have a ground truth object because they are on the background $c_j^{GT}=0$ and  $s_j^{GT}=0$ and the localization loss will be zero.
The smooth $L1$ distance is used as a measure of the error
\begin{eqnarray*}
\text{smooth}_{L1}(x) = \left\{\begin{array}{ll}
0.5x^2 & \text{if }|x| < 1, \\
|x| - 0.5 & \text{otherwise.}
\end{array}\right.
\end{eqnarray*}



There exists some related work which directly aims to regress to
part (sub-object) coordinates. Liang et al.~\cite{liang2015deep} first
detect persons and then train a network to regress to the parts
(clothes) given the detected person. Other than our method they
separate the object detection and part detection. Cervantes et
al.~\cite{cervantes2016} propose a method which regresses directly to
the object and parts. However, their approach only applies to images
where a single object is present in the image.

Compared to the Recursive Fast-CNN of Section~\ref{sec:fast-rcnn}, the
Multi-stage model we propose here allows us to skip one stage and
therefore inspect fewer pixels and reduce bandwidth
consumption. However, this could potentially come at the cost of
performance loss because the regression to the sub-object bounding box
coordinates is done from the low resolution input image.

\section{Experimental results}
\label{sec:experiments}

\begin{figure}
\begin{center}
\setlength{\tabcolsep}{0.1em}
\begin{tabular}{ccc}
\subfloat{\label{fig:difficulty1}\includegraphics[width = 0.3\linewidth]{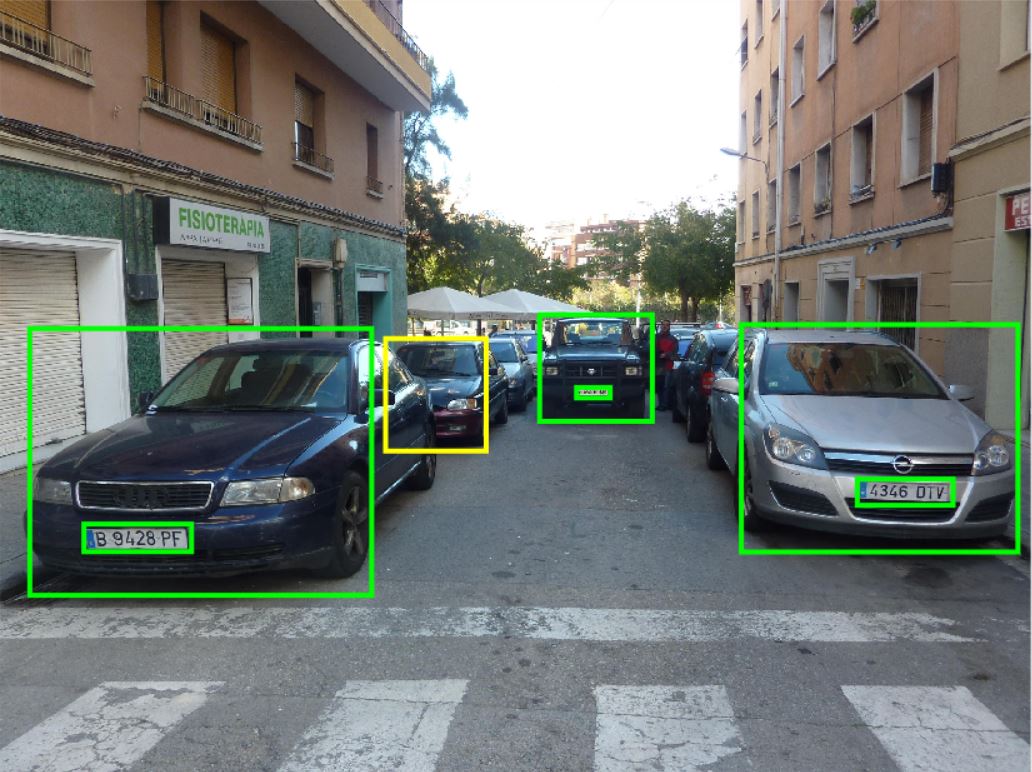}}&
\subfloat{\label{fig:difficulty2}\includegraphics[width = 0.3\linewidth]{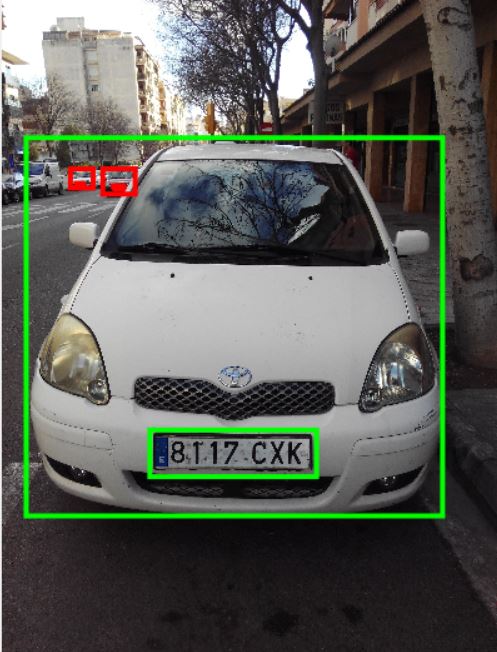}} &
\subfloat{\label{fig:difficulty3}\includegraphics[width = 0.3\linewidth]{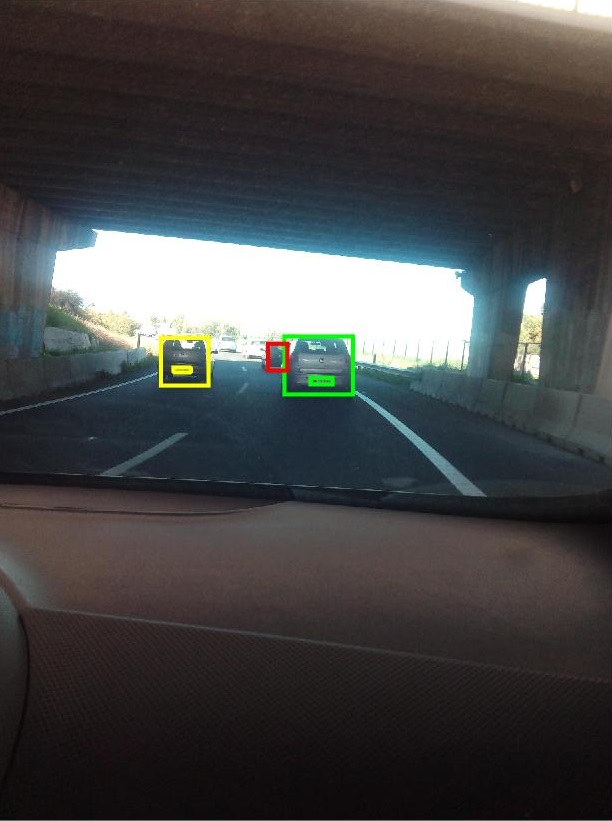}}
\end{tabular}
\caption{Example images from our car recognition dataset with the
  corresponding ground truth: green are ``Easy'' detections, yellow
  ``Medium'' and red ``Hard''.}
\label{fig:examples_difficulty}
\end{center}
\end{figure}

In this section we report on a series of experiments performed to
quantify the performance of our approach in terms of recognition
accuracy and bandwidth savings. We first introduce a new car
recognition dataset we collected to evaluate our two models.

\subsection{A dataset for car recognition}
\label{sec:dataset}


Car datasets often cannot be used for license
plate detection and recognition because the average car resolution is
too low, while license plate datasets usually do not contain the full
car. Consequently, most car or license plate datasets focus on one of the two problems,
and rarely on both. Because of this, we introduce a new dataset for benchmarking car
detection, license plate detection, and license plate
recognition.\footnote{The dataset will be released upon publication of
  this work.}

The dataset consists of 500 images taken on different days, with
different devices, at different times and in different cities. All
images have been manually annotated with bounding box coordinates of
the car and the license plate, and the characters of the license plate
were transcribed (when legible). Due to the variability in difficulty,
all the annotations include a measure of its difficulty as ``Easy'',
``Medium'' or ``Hard''. This difficulty was determined considering
occlusions, size and definition of the object. In
Figure~\ref{fig:examples_difficulty} we show some examples of images
from the dataset with their corresponding difficulty annotation.

\begin{figure*}[t]
\begin{tabular}{cc}
\setlength{\tabcolsep}{0.1em}
\includegraphics[width = 0.5\linewidth, height=4cm]{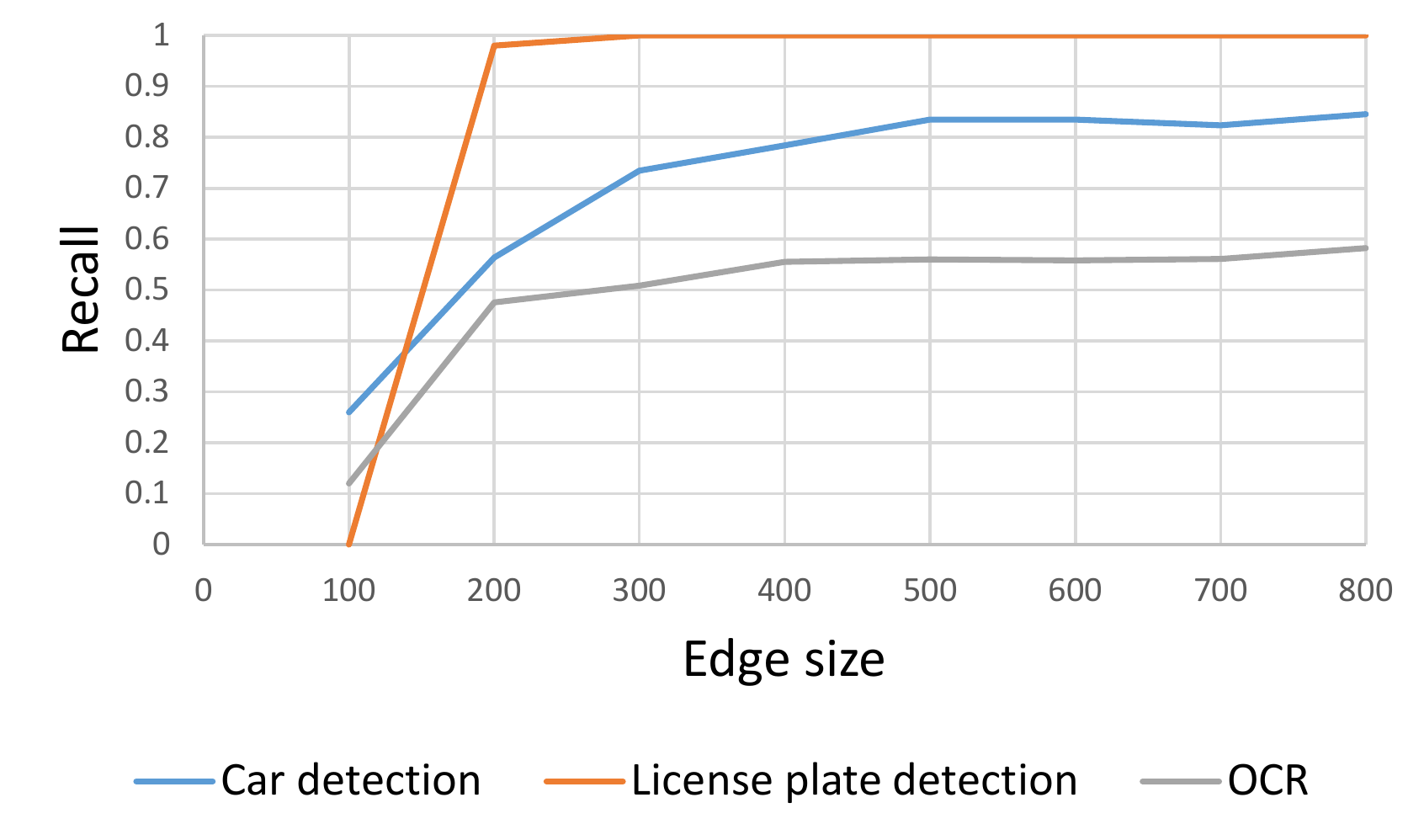} &
\includegraphics[width = 0.5\linewidth, height=4cm]{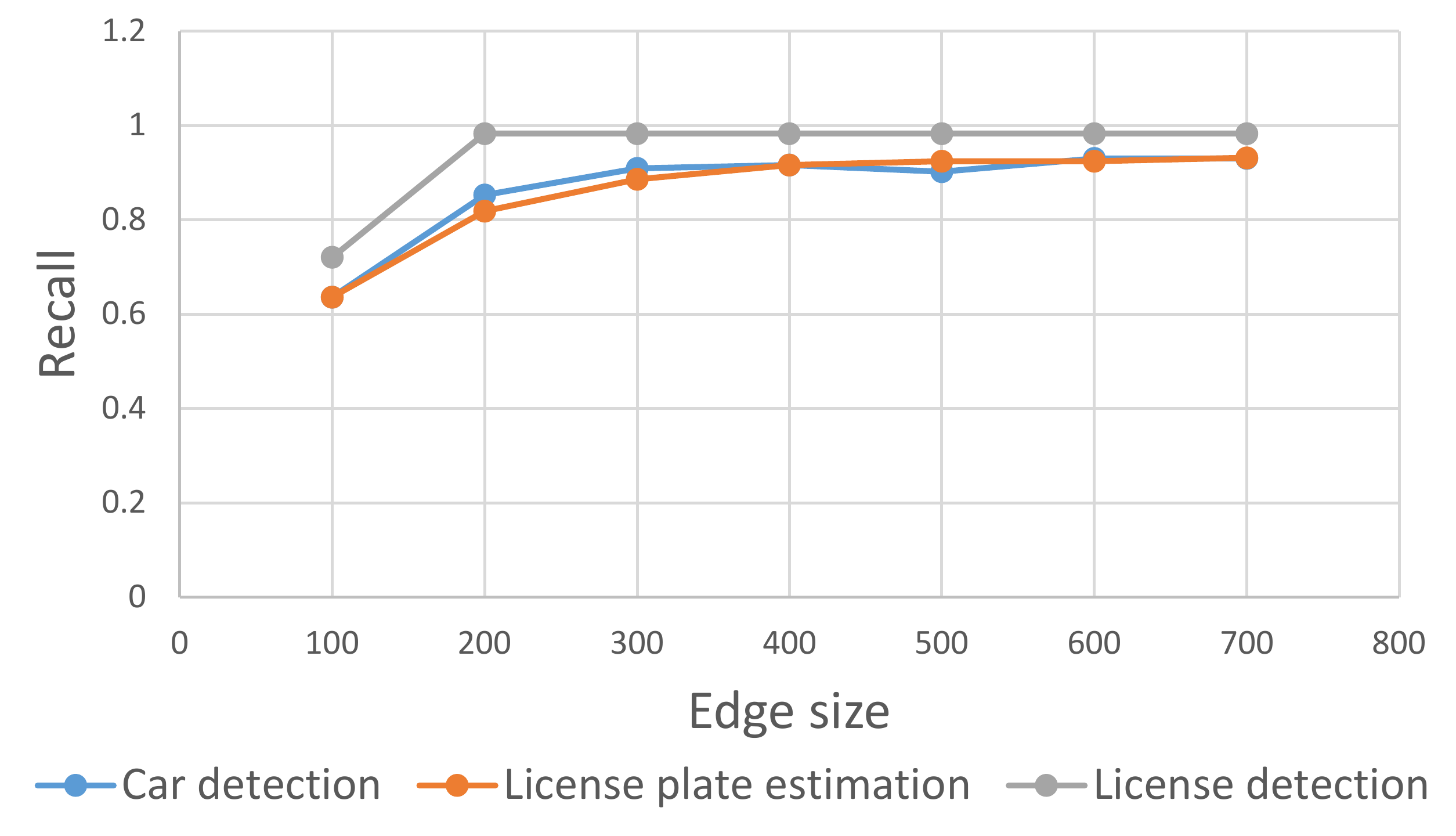} \\
(a) Recursive Fast-RCNN & (b) Multi-stage Network
\end{tabular}
\caption{Experimental results. (a) Car and license plate recall and
  character accuracy on the test fold with respect to the size of the
  image (in number of pixels on the longest edge) for the model 2. (b)
}
\label{fig:parameter_cross_validation}

\end{figure*}

\subsection{Car recognition in high resolution imagery}

We consider the task of car recognition in high resolution imagery as
a practical application of the framework that we propose. By car
\emph{recognition} we mean recognition of its license plate since it
is a unique way of identifying a car. Therefore, the final goal of the
system is to read the characters of the license plate of all cars
present in an image. License plate characters are normally relatively
small with respect to the whole image and therefore not visible in a
low resolution version of the image (see again
Figure~\ref{fig:introduction}). However, since we know that the
characters are going to be inside a license plate and the license
plate at the same time will be contained in a car we can apply the
models we proposed in Section~\ref{sec:approach}. Both systems were
trained on 75\% of the images from the dataset proposed in
Section~\ref{sec:dataset} and tested on the remaining 25\%.

For the Recursive Fast-RCNN model, we trained a car detector by
finetuning a Fast-RCNN detector based on the VGG\_CNN\_M\_1024
network\footnote{https://github.com/rbgirshick/fast-rcnn}. The license
plate detector for the second Fast-RCNN network was trained on the
regions of the training set containing cars. When testing the system,
we pass the original images at low resolution to the car detector and
select detections with a score of 0.3 or higher for inspection at
higher resolutions by the license plate detector. Since we know that a
car only contains one license plate, instead of considering all the
license plate detections with a 0.3 score or higher we only consider
the detection with highest score greater than 0.3. Finally, the region
detected as a license plate is requested at higher resolution and
passed to the Tesseract OCR engine~\cite{smith2007overview}. More
details about the resolutions used at each are given in
Section~\ref{sec:baseline}.

For the Multi-stage model, we trained the network proposed in
Section~\ref{sec:multi-stage} with a car detector and a license plate
estimator. We consider all car detections with its corresponding
license plate estimation with a score of 0.3 or higher. License plates
are very small objects with respect to the entire image, and
especially with respect to the resolution of the final convolutional
layer in state-of-the-art networks. Consequently, estimation of the
license plate position at such low resolution is not very precise and
we add some margin to the estimation in order to ensure that the
license plate is contained in that region. The margin $M_g$ added
to the estimation is proportional to the estimated probability
${p}(s | \mathbf{\hat{b}}_{i|s})$ that the predicted region contains the
license plate. Indeed, we define $M_g$ as follows:
\begin{eqnarray*}
M_g = 1 - {p}(s | \mathbf{\hat{b}}_{i|s}).
\end{eqnarray*}
The final predicted bounding box is computed by adding this margin to
the predicted bounding box from the network, constraining it to be
contained in the object bounding box from the previous level. Then, a
license plate detector similar to the one used for the first model is
run on the prediction of the license plate location. Finally, the
license plate region is passed to the Tesseract OCR at high
resolution for the final recognition.

\subsection{Experimental evaluation}
\label{sec:baseline}

An important parameter is the optimal resolution at which the
successive images should be processed. For the Recursive Fast-RCNN
model we must decide the resolution of the original image at which the
car is detected, the resolution at which to request the car to
perform license plate detection, and finally the resolution at which
to request the license plate before passing it to Tesseract. For the
Multi-stage Network model, instead of determining the resolution of
the car we have to request the estimation of the license plate
localization.

In order to determine these parameters we divided the training set into
two folds; we used 75\% of the training images to train the model and
the other 25\% of images for parameter crossvalidation. Considering the Recursive Fast-RCNN model, in
Figure~\ref{fig:parameter_cross_validation}(a) we show the recall of
car and license plate detection and character accuracy for different
image resolutions. Car detection recall keeps increasing up to 500
pixels and then stabilizes. For license plate detection, there is a
clear peak at 300 pixels on the longest edge of the car image, then
recall stabilizes. Based on these results, we set the resolution of
the original image to 500 pixels and of the car image to 300
pixels. We found OCR performance to keep increasing with resolution
and therefore we use all available resolution for character
recognition. Taking now into account the Multi-stage network model, in Figure~\ref{fig:parameter_cross_validation}(b) we show
similar experiments where we determine to use 600 pixels on the
longest edge for the original resolution in order to have a good car
detection and license plate regression and 200 pixels on the longest
edge for the license plate location estimation.

\begin{figure}[t]
\begin{center}
\includegraphics[width=0.9\linewidth]{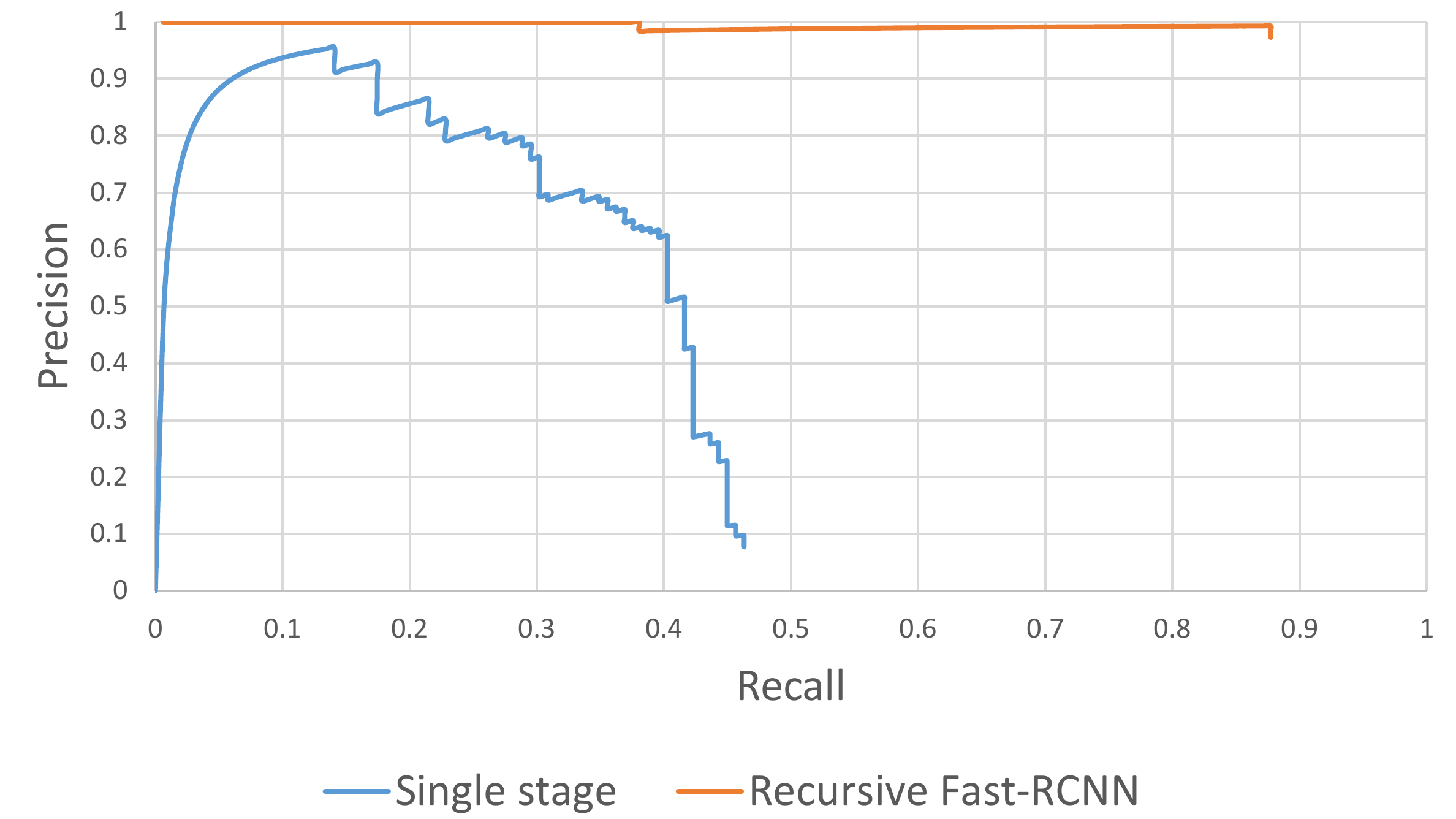}
\caption{Comparison between our Multi-stage model
  and a single-stage approach for license plate detection.}
\label{fig:comp_sing_multi}
\end{center}
\end{figure}

\begin{figure}
\begin{center}
\includegraphics[width = 0.9\linewidth, height=4cm]{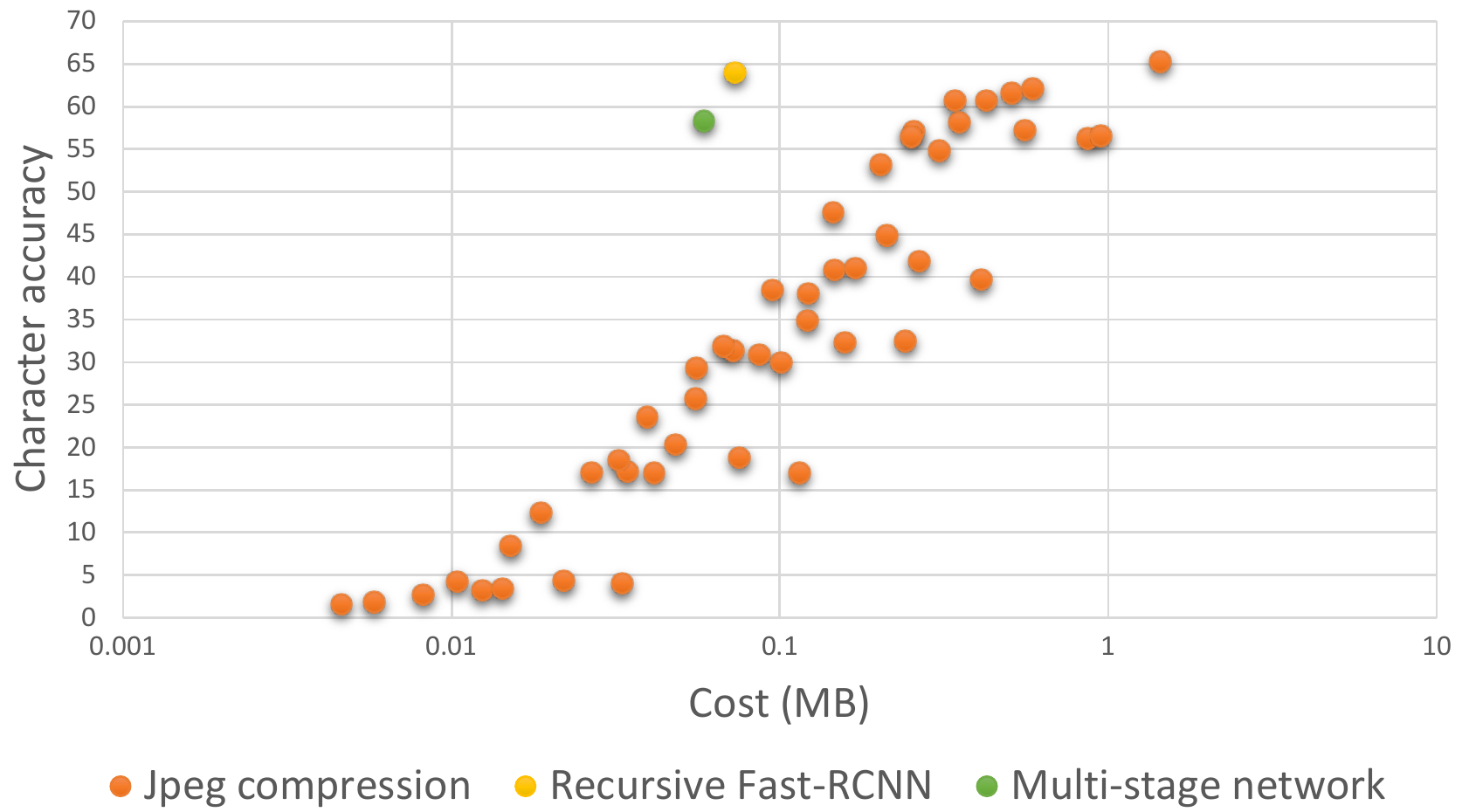}
\caption{Comparison with JPEG compression in terms of character accuracy and
  bandwidth cost (in log-megabytes).}
\label{fig:jpeg_comparison}
\end{center}
\end{figure}

\begin{figure}
\begin{center}
\includegraphics[width = 0.75\linewidth]{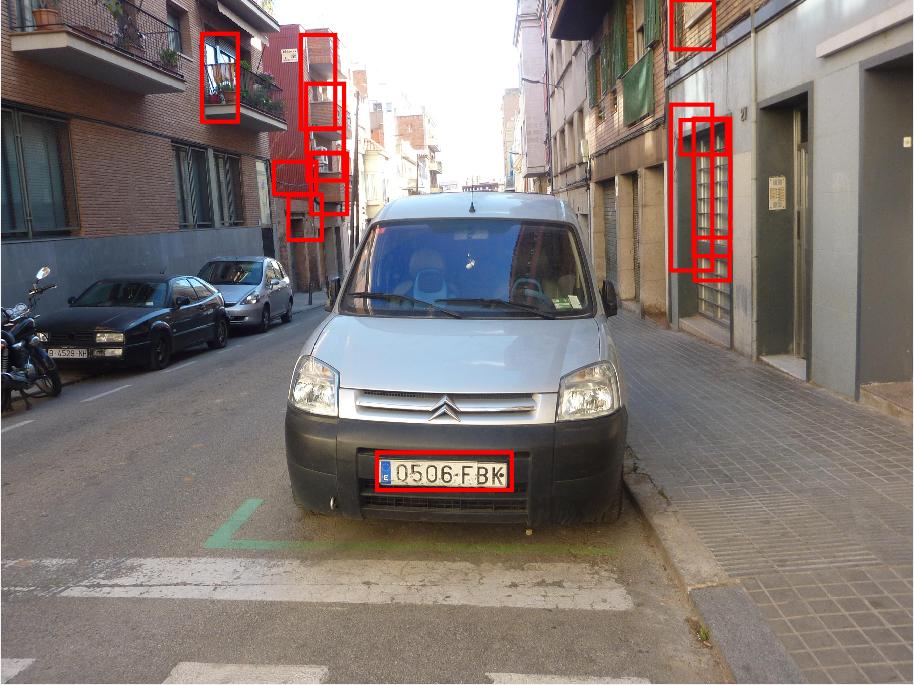}
\caption{Output from a network trained to detect license plates in the original image without previously detecting the car.}
\label{fig:single_stage}
\end{center}
\end{figure}

Our proposed framework is not only useful for optimizing bandwidth, but also to increase
performance. This is shown in Figure~\ref{fig:comp_sing_multi}, where
we demonstrate the improvement in performance that the Recursive Fast-RCNN
model yields with respect to training a neural network to directly
search for license plates. The single stage approach has a
considerably lower precision-recall curve. The difference is
remarkable; using an active seeking approach prevents detection of
false-positives where no cars exist, and knowing the car position
provides important information on plausible license plate locations. In Figure \ref{fig:single_stage} we show an example of the output from a network trained to directly detect license plates without previously detecting the car.




\subsection{Baseline comparison with image compression}

\begin{figure*}[ht]
\centering
\begin{tabular}{cc}
\includegraphics[width=0.45\linewidth]{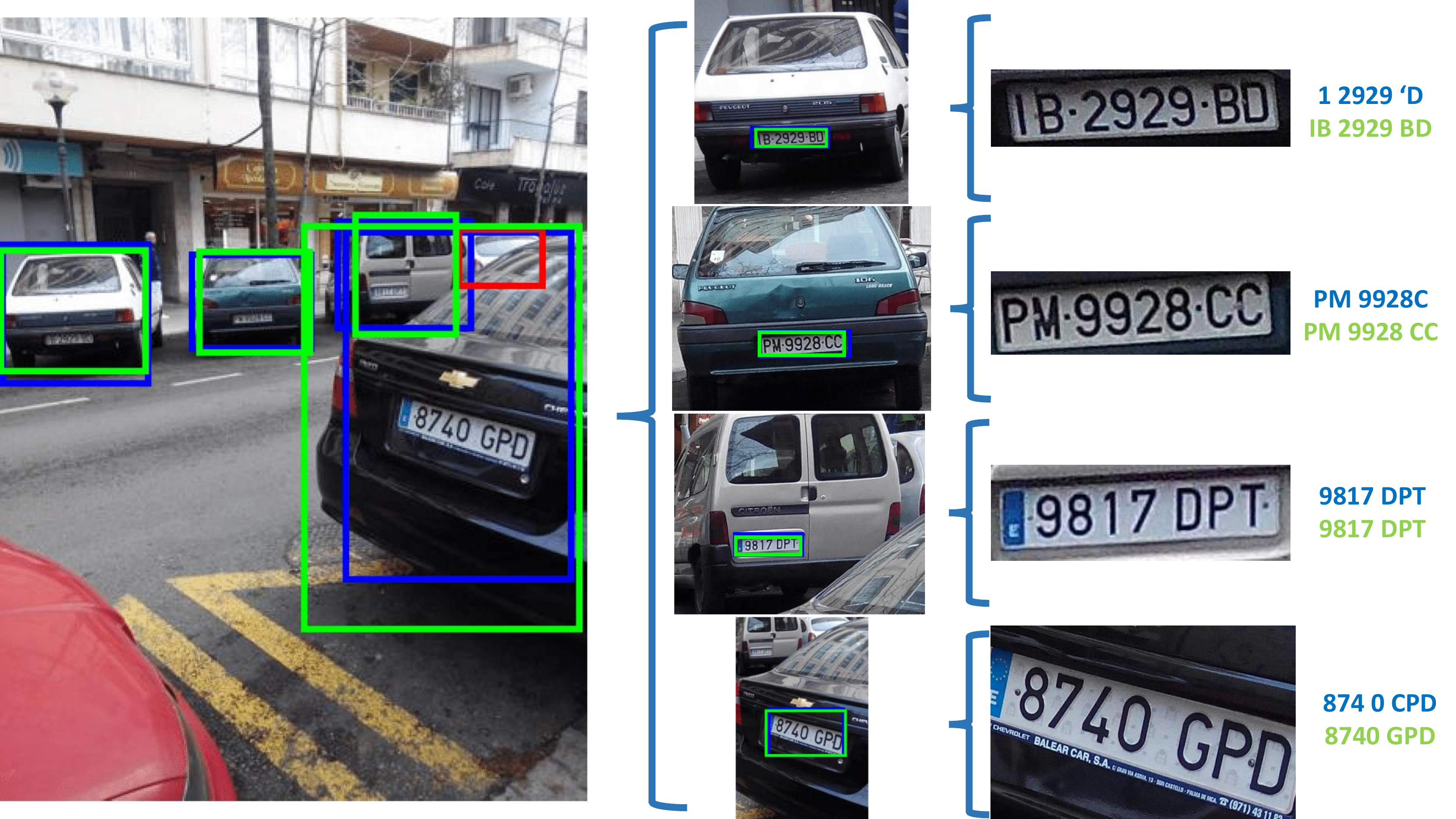} &
\includegraphics[width=0.45\linewidth]{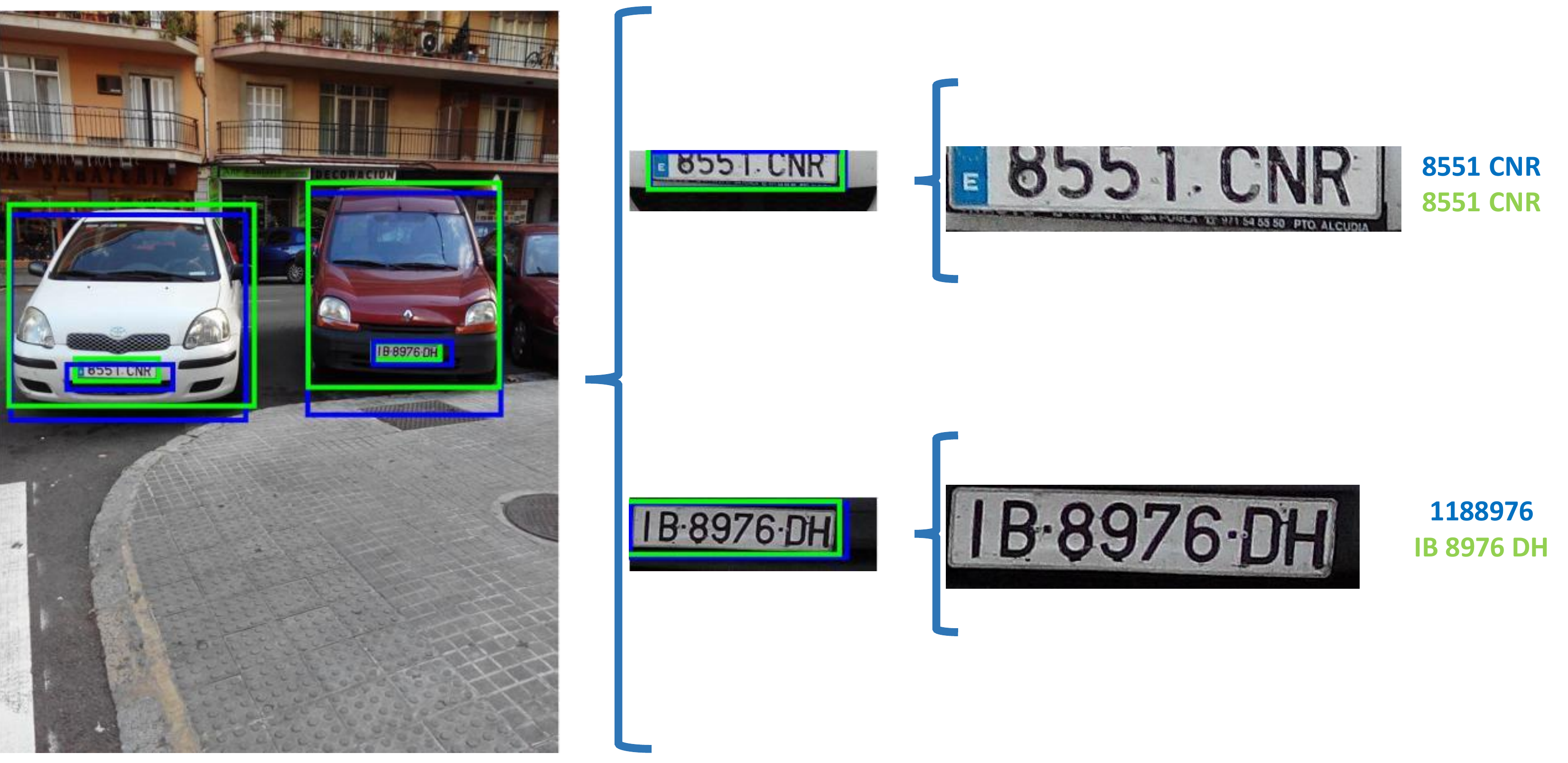} \\
(a) Recursive Fast-RCNN & (b) Multi-stage Network
\end{tabular}
\caption{Qualitative results from both models. Blue represents the output given
  by the algorithm on the different steps, green represents ground
  truth with ``Easy'' difficulty and red ground truth with ``Hard''
  difficulty.}
\label{fig:qualitative_result}
\end{figure*}

The objective of our paper is to use semantic classes to reduce the
bandwidth required for recognition. This is a type of \emph{semantic
  image compression}, and to the best of our knowledge there is no
prior work on task-specific image compression. As a baseline we
consider a system that compresses using JPEG and/or image resizing.
Both operations reduce the bandwidth required to transmit the
data. However, they also may negatively impact OCR character
recognition performance. Our baseline system based on image
compression is endowed with perfect license plate detection (i.e. we
use the groundtruth license plate boxes). We evaluate OCR performance
for five different resolutions $R=\{500,1000, 1500, 2000,3000,4000\}$,
and eight different levels of JPEG compression quality
$Q=\{1, 5, 10, 15, 20, 25, 50, 75\}$. This leads to 40 possible pairs
of image resolution and compression quality whose performance as a
function of the cost (in log-megabytes, see
Section~\ref{sec:method}) on the test set is plotted as orange points
in Figure~\ref{fig:jpeg_comparison}. The results of the Recursive
Fast-RCNN are plotted as a yellow dot and the Multi-stage network as a
green one.

As shown in Figure~\ref{fig:jpeg_comparison}, both the Recursive
Fast-RCNN and the Multi-stage network yield significant savings in
terms of cost when compared to the JPEG model with an equivalent
accuracy. The Recursive Fast-RCNN model reduces the cost over one
order of magnitude and still reaches 64\% character accuracy which is
only 1\% lower than the 65\% character accuracy obtained by the JPEG
baseline when sending the image at its original resolution. The
Multi-stage Network is able to decrease the cost by 20\% with respect
to Recursive Fast-RCNN from an average of 0.073MB to 0.58MB. Pixel-wise, the Recursive Fast-RCNN framework requires the inspection an average of 18.6\% of the pixels while the Multi-stage Network inspects an average of 17.6\% of the pixels. This comes at a cost of 5\% loss in character accuracy. This drop in
accuracy is caused by the low resolution at which we are detecting
small details from the image. However, we believe that the network
could be improved by training over a bigger dataset.

When making the comparison between the proposed framework and the JPEG
baseline, we stress that we have given the JPEG baseline algorithm the
advantage of \emph{perfect car and license plate detection}, while our
framework must both \emph{detect and localize} the license plate
before recognition.

Figure~\ref{fig:qualitative_result} qualitatively illustrates the step
by step execution of both proposed algorithms. Note that all the ``Easy''
cars and license plates are correctly detected (only one ``Hard'' car
instance is missed).


\section{Conclusions and future work}
\label{sec:conclusions}
In this paper we proposed a framework for identifying promising
regions in low resolution images and progressively requesting regions at higher resolution to perform recognition of a higher semantic quality. This framework is especially interesting to reduce the bandwidth needed in order to take advantage of all available resolution of modern high resolution cameras when recognizing objects. We propose two different models implementing license plate recognition within this framework. The first implementation consists of a Recursive Fast-RCNN model which uses a Fast-RCNN network at each level of the framework to perform object detection. The second model uses a Multi-stage Network to simultaneously localize and recognize both cars and their license plates at the first level of resolution, and thus allowing the network to inspect fewer pixels at subsequent levels. 

Experimental results on both implementations show that the proposed framework yields significant savings in terms of bandwidth cost (measured with the number of pixels inspected for license plate recognition) when compared JPEG compression at an equivalent plate recognition accuracy.

The framework we propose is not incompatible with JPEG compression so as future work we would like to combine this framework with different JPEG compression levels to improve even further the savings in terms of bandwidth. Moreover, although the implementation of an OCR was not in the scope of this work, we would like to investigate in this field to improve the character accuracy.
\section*{Acknowledgement}
This work is funded by the Projects TIN2013-41751-P, TIN2016-79717-R, TIN2014-52072-P and TIN2016-75404-P of the Spanish Ministry of Economy, Industry and Competitiveness with FEDER funds and the Chist-ERA project M2CR (PCIN-2015-251) funded by MINECO through APCIN 2015, the Catalan project 2014 SGR 221, and the CERCA Programme. We also thank NVIDIA for the generous GPU donation. Laura Lopez-Fuentes benefits from the NAERINGSPHD fellowship of the Norwegian Research Council under the collaboration agreement Ref.3114 with the UIB.

{\small
\bibliographystyle{ieee}

}

\end{document}